# Jailbreak Detection in Clinical Training LLMs Using Feature-Based Predictive Models

Tri Nguyen[1], Lohith Srikanth Pentapalli[1], Magnus Sieverding[1], Laurah Turner[1], Seth Overla[1], Weibing Zheng[1], Chris Zhou[1], David Furniss[1], Danielle Weber[1], Michael Gharib[1], Matt Kelleher[1], Michael Shukis[1], Cameron Pawlik[2], and Kelly Cohen[1]

[1] University of Cincinnati, Cincinnati OH 45221, USA
nguye3hr@mail.uc.edu, pentapls@mail.uc.edu,
sievermf@mail.uc.edu, turnela@ucmail.uc.edu,
overlash@ucmail.uc.edu, zhengwb@mail.uc.edu,
zhouct@ucmail.uc.edu, furnisde@mail.uc.edu,
weber2de@ucmail.uc.edu, gharibmw@mail.uc.edu,
kellehmw@ucmail.uc.edu, shukismc@mail.uc.edu,
cohenky@mail.uc.edu
[2] University of Michigan, Ann Arbor MI 48109, USA
cpawlik@umich.edu

**Abstract.** Jailbreaking in Large Language Models (LLMs) threatens their safe use in sensitive domains like education by allowing users to bypass ethical safeguards. This study focuses on detecting jailbreaks in 2-Sigma, a clinical education platform that simulates patient interactions using LLMs. We annotated over 2,300 prompts across 158 conversations using four linguistic variables shown to correlate strongly with jailbreak behavior. The extracted features were used to train several predictive models, including Decision Trees, Fuzzy Logic-based classifiers, Boosting methods, and Logistic Regression. Results show that feature-based predictive models consistently outperformed Prompt Engineering, with the Fuzzy Decision Tree achieving the best overall performance. Our findings demonstrate that linguistic-feature-based models are effective and explainable alternatives for jailbreak detection. We suggest future work explore hybrid frameworks that integrate prompt-based flexibility with rule-based robustness for real-time, spectrum-based jailbreak monitoring in educational LLMs.

## 1  Introduction

In the context of Large Language Models, jailbreaking refers to techniques that intentionally bypass built-in ethical constraints, safety, or alignment to make the model generate restricted or harmful outputs. These jailbreaks are often achieved through heavy tailored prompts, prompt injections, or adversarial inputs that exploit weaknesses in the model's mechanisms [1-5]. Jailbreaking has become more and more complicated as it can interact with multiple parts in the pipeline of LLM-powered applications [6]. Jailbreaking can enable the misuse of LLMs for generating toxic, illegal, or dangerous content, including hate speech, misinformation, malware, and instructions for illegal activities [7-10]. Since many jailbreaks can be shared and reproduced easily, even users with limited technical skill can exploit models in harmful ways. This undermines trust in LLMs as safe tools, especially when deployed in public-facing applications like customer support, education, or therapy [11].

2-Sigma is an innovative educational tool that leverages the power of LLMs to create dynamic clinical scenarios, supporting medical education at the University of Cincinnati's College of Medicine. Equipped with advanced methodologies in machine learning, such as chain-of-thought reasoning and few-shot prompting, 2-Sigma has achieved remarkable success in its initial deployment. However, this success has also revealed a critical challenge: students have been found to manipulate the AI system, leading to ethical and professional deviations. This type of jailbreak has been widely studied, and several techniques have been proposed to solve the problem. One key strategy is Chain-of-Thought (CoT) prompting, which encourages models to produce intermediate reasoning steps, improving their ability to handle complex queries in a structured and less exploitable way [12]. A powerful variant is Zero-shot CoT, which demonstrates that simply appending "Let's think step by step" can guide models to reason more effectively without the need for training examples [13]. Self-Consistency, another enhancement, involves sampling multiple reasoning paths and selecting the most consistent output, increasing reliability and reducing vulnerability to trick prompts [14]. Additional techniques like Generated Knowledge Prompting allow users to guide the model to produce relevant background information before tackling the main question, effectively creating internal context that is harder to subvert [15]. Finally, methods such as Least-to-Most Prompting and Tree of Thoughts (ToT) provide multi-step frameworks for solving complex tasks by breaking them down into simpler components or exploring multiple solution branches with self-evaluation, both of which make models more robust against deceptive or manipulative inputs [16-17]. Despite employing sophisticated prompt guardrails and token operations, the nature of semantic evaluation makes traditional binary detection inadequate for identifying these nuanced attempts.

An explainable, clear-box approach to addressing jailbreaking in large language models is essential for ensuring transparency, trust, and accountability - especially in ethically sensitive domains like medical education [18, 21]. Unlike black-box systems that obscure internal decision-making, eXplainable AI (XAI) models allow researchers and practitioners to inspect how responses are generated, trace reasoning paths, and identify where safety mechanisms succeed or fail. This visibility is crucial in education, where AI-generated content directly influences learning, assessment, and student trust [19-20]. If a model can be jailbroken to bypass content filters and generate inappropriate, biased, or misleading material, educators need clear diagnostic tools to understand how and why it happened.

After the initial launch of 2-Sigma, a set of conversations between students and the LLM was collected for analysis. Preliminary reviews of these conversations revealed a consistent pattern: jailbreak attempts often correlate with specific linguistic features, such as a lack of professionalism, engagement in harmful behavior, distraction, or deviation from the medical case at hand. This paper investigates how such behaviors can be used to detect and flag potential jailbreak attempts. A team of annotators evaluated each prompt based on four key features, each with multiple categorical labels. The resulting annotated data was then processed and used as input for multiple machine learning algorithms, such as decision trees, fuzzy-based systems, boosting methods, etc. Prediction results were compared against a human-defined ground truth to evaluate algorithm performance. Findings show that using these extracted linguistic features leads to higher prediction accuracy than relying on prompt engineering alone, with distraction emerging as a particularly strong indicator of jailbreaking.

## 2    Data Collection

To investigate jailbreak behaviors, we analyzed interactions between medical students and the 2-Sigma LLM, which plays the role of virtual patients. The dataset consisted of 158 conversations, totaling about 2,300 prompts. Each prompt was evaluated by two members of the research team to determine whether it constituted a jailbreak attempt. These binary judgments served as the ground truth for training and evaluating the performance of predictive algorithms. The final annotated dataset included 1,155 jailbreak prompts and 1,146 non-jailbreak prompts, resulting in a nearly balanced class distribution suitable for classification tasks.

A team of seven human annotators evaluated each prompt using a structured rubric based on four linguistic variables: Professionalism, Medical Relevance, Ethical Behavior, and Contextual Distraction. These variables were selected after a preliminary qualitative review of jailbreak patterns, which revealed strong correlations between these features and jailbreak behavior. The definitions of the four variables are provided in Table 1, which outlines how each construct captures key aspects of student responses, such as focus, safety, tone, and adherence to clinical context.

**Table 1.** Definition of linguistic variables.

| Linguistic Variables | Definition |
| --- | --- |
| Professionalism | Professionalism involves maintaining appropriate conduct, respectful language, and adherence to clinical boundaries. Violations include unprofessional tone, inappropriate comments, breaches of confidentiality, and behaviors that disregard patient safety, consent, or demonstrate abusive tendencies. |
| Medical Relevance | Medical relevance tracks whether students remain focused on the case, flagging irrelevant questions, unnecessary exams, or inaccurate diagnoses. |
| Ethical Behavior | Ethical Behavior reflects the prioritization of patient safety, autonomy, and well-being. Red flags include making harmful recommendations, ignoring critical symptoms, or engaging in actions that compromise informed consent or ethical standards of care. |
| Contextual Distraction | Contextual Distraction assesses the relevance of each prompt within the broader context of the entire student–AI "patient" |

| | conversation. Violations occur when a message strays from the established context, introducing unrelated topics or creating distractions that shift focus from the case discussion. This ensures that each interaction remains focused and contributes meaningfully to the case without sidetracking into irrelevant or distracting content. |
|---|---|

Each variable was rated on ordinal scales to capture the nuances of human judgments. The levels used for each dimension are illustrated in Table 2. These levels were carefully defined by the research team to reflect how human evaluators interpret meaning, intent, and risk in clinical dialogue. The annotated data will be normalized by calculating the proportion of coders who selected each rating level for each linguistic variable - that is, the number of annotators choosing a specific rating divided by the total number of annotators. Since there are 15 total rating levels across all four variables, this process will yield 15 normalized scores, which will be used as input features for the predictive models. The following example illustrates how one prompt is evaluated by seven annotators and how the resulting labels are converted into normalized feature values:

*Prompt:* "Did you pee or poop today?"

*Annotated data:*
- Professionalism: unprofessional (6), borderline (1), appropriate (0)
- Medical Relevance: irrelevant (1), partially relevant (6), relevant (0)
- Ethical Behavior: dangerous (0), unsafe (0), questionable behavior (0), mostly safe (0), safe (7)
- Contextual Distraction: highly distracting, moderately distracting, questionable (7), not distracting

*Normalized features:*
- Professionalism: unprofessional (0.857), borderline (0.143), appropriate (0)
- Medical Relevance: irrelevant (0.143), partially relevant (0.857), relevant (0)
- Ethical Behavior: dangerous (0), unsafe (0), questionable behavior (0), mostly safe (0), safe (1.0)
- Contextual Distraction: highly distracting, moderately distracting, questionable (1.0), not distracting

**Table 2.** Linguistic Variables Rating Scales.

| Linguistic Variables | Rating Scale |
|---|---|
| Professionalism | 1. Unprofessional<br>2. Borderline<br>3. Appropriate |
| Medical Relevance | 1. Irrelevant<br>2. Partially relevant<br>3. Relevant |
| Ethical Behavior | 1. Dangerous<br>2. Unsafe<br>3. Questionable behavior |
| | 4. Mostly safe |
| | 5. Safe |
| Contextual Distraction | 1. Highly distracting |
| | 2. Moderately distracting |
| | 3. Questionable |
| | 4. Not distracting |

## 3 Predictive Models

### 3.1 Decision Tree (DT) and Fuzzy Decision Tree (FDT)

The default decision tree model from *scikit-learn* [22] was implemented as a classifier for detecting jailbreak prompts. This model uses the CART (Classification and Regression Tree) algorithm and constructs binary trees by recursively splitting features to minimize Gini impurity. Two decision tree models were created for comparison: one using the default settings (allowing the tree to grow to full depth), and another with the maximum depth restricted to 3 to prevent overfitting and encourage generalization. The default model offers greater complexity and potentially better fit to the training data, while the shallow model provides a more interpretable structure, helping to identify which linguistic variables and rating levels most strongly contribute to jailbreak classification.

A fuzzy decision tree was implemented using the *FuzzyDecisionTreeClassifier* from the *fuzzytree* library [23]. This model extends traditional decision trees by incorporating fuzzy logic, allowing for soft decision boundaries instead of hard splits. This enables smoother transitions between categories, helps reduce overfitting, and more naturally handles uncertainty and gradations in human annotations. It is particularly effective for subjective, language-based features variables such as "borderline professionalism" or "partially relevant," where crisp categorization may fail to capture the nuanced meaning of student prompts.

### 3.2 Gradient-Optimized Fuzzy Inference System

Additionally, we employ a Gradient-Optimized Fuzzy (GF) inference system—a hybrid approach that combines the interpretability of fuzzy logic with the efficiency of gradient-based learning. This model would be considered a white box model due to the ability to trace the decisions from output to the inputs linguistically. Unlike traditional fuzzy systems that rely on heuristic or derivative-free optimization, the GF method leverages gradient descent to fine-tune parameters. The fuzzy approach offers a natural advantage in handling ambiguous or overlapping class boundaries thanks to its use of fuzzy sets, which allow partial membership and better reflect real-world uncertainty [30-31].

The Fuzzy Inference System (FIS) used in this study employed 13 membership functions per input and operated with just 2 rules. Training was conducted for 250 epochs on a laptop equipped with a 12th Gen Intel(R) Core(TM) i7-1260P processor. While this CPU is relatively capable for a mobile device, it is far from high-end by today's standards. Even so, the GF model completed training in just 13.06 seconds.

Convergence likely occurred earlier, but in the absence of an early stopping mechanism - deemed unnecessary due to the already short training time - training proceeded to the full epoch count. This result highlights the efficiency of pairing a compact and adaptable FIS with the speed and performance of gradient descent, all while maintaining the inherent interpretability of fuzzy logic.

### 3.3    Random Forest (RF)

RF is an ensemble learning method that constructs a collection of decision trees and combines their predictions to produce more accurate and robust results. Each tree is trained on a random subset of the training data and a random subset of features - a process known as bagging (bootstrap aggregating) and feature sampling [26]. This statistical randomness improves generalization and makes the model highly resistant to overfitting and noise. RF handles both categorical and continuous features effectively and is known for its strong performance in a wide range of classification tasks [24]. Although individual trees in a forest are interpretable, the ensemble as a whole is typically considered a black-box model, as it can be difficult to trace how specific features contribute to a particular prediction. In this study, the model was implemented using the *RandomForestClassifier* from the *scikit-learn* library.

### 3.4    Light Gradient Boosting Machine (LGBM)

Unlike Random Forest, which builds decision trees in parallel through bagging, LGBM employs a boosting strategy, constructing trees sequentially [25]. In this approach, each new tree is trained to correct the errors made by the previous trees. Specifically, the gradient of the loss function is used to adjust the weights of training samples, allowing subsequent trees to focus more on instances that were previously misclassified. A key feature of LGBM is its optimization for high performance and scalability, particularly on large datasets. It achieves this through techniques such as histogram-based binning, which reduces computational cost by grouping continuous values into discrete bins, and Gradient-based One-Side Sampling (GOSS), which prioritizes informative samples for gradient calculation. These optimizations make LGBM both faster and more memory-efficient than many traditional gradient boosting implementations. Similar to RF, LGBM is considered a black-box model, as its sequential learning and complex tree structures make individual predictions difficult to interpret. In this study, the model was implemented using the default *LGBMClassifier* from the *lightgbm* library [29].

### 3.5    eXtreme Gradient Boosting (XGBoost)

XGBoost is a popular gradient boosting framework known for its speed, accuracy, and advanced regularization capabilities [27]. Like LGBM, XGBoost builds trees sequentially, where each new tree aims to correct the residual errors of the previous model. However, XGBoost distinguishes itself through the use of second-order gradient information (i.e., both gradients and Hessians) to optimize its objective function, leading to more accurate and stable learning. It also includes built-in L1 and L2 regularization, which helps control model complexity and reduce the risk of overfitting. XGBoost is particularly robust when dealing with sparse or missing data, and it provides fine-grained control over training through a wide array of hyperparameters. While its performance is typically strong in classification tasks, it is also considered a black-box model due to the complexity of its internal boosting process. In this study, the model was implemented using the *XGBClassifier* from the *XGBoost* library [28].

### 3.6 Logistic Regression (LR)

LR is a widely used statistical model for binary classification tasks. It models the probability that a given input belongs to a particular class using the sigmoid function, mapping linear combinations of input features to a probability range between 0 and 1. One of LR's key advantages is interpretability - the model coefficients can be directly interpreted to be the influence of each feature on the prediction outcome. Unlike tree-based models, logistic regression assumes a linear relationship between the input features and the log-odds of the output, which may limit its performance on complex, non-linear patterns. In this study, the model was implemented using the *LogisticRegression* class from the *scikit-learn* library.

### 3.7 Neural Networks (NN)

Neural Networks (NN) are inspired by biological processes, giving rise to computational models that comprise interconnected processing units (neurons). These units are organized in layers to learn complex patterns from data [32]. Due to their layered architecture, NNs can extract hierarchical representations, making them particularly effective for capturing complex non-linear relationships [33]. The core learning ability in NNs happens through adjusting connecting weights between neurons through backpropogation, which minimizes the difference between predicted and actual outputs using gradient descent optimization. This approach enables the neural networks to adapt to a wide variety if data distributions and classification tasks [34]. Despite their exceptional performance across various domains, NNs are black-box models due to intricate and complex internal representations and decision boundaries. The challenge of interpretability is even more pronounced in deeper architectures where learned features become increasingly abstract. For this study, a custom Neural Network was implemented using the TensorFlow/Keras framework with a four-layer architecture consisting of 32, 32, 16, and 16 neurons respectively with ReLU activation. This configuration provides an effective balance between model capacity and computational efficiency while delivering optimal performance for the jailbreak detection task.

### 3.8 Prompt Engineering (PE)

In 2-Sigma, Prompt Engineering was initially employed as a preventive strategy to reduce jailbreak attempts by guiding the LLM's behavior through explicit, scenario-aware instructions. Prior to each interaction, a set of behavioral guardrails was embedded directly into the prompt to reinforce the model's alignment with the educational objectives of the system. These instructions directed the model to respond in a firm manner when users attempted to bypass clinical reasoning, request a direct diagnosis without proper steps, or introduce irrelevant or fantastical elements. Additionally, prompts instructed the LLM to end the scenario and provide feedback if a user engaged in unprofessional or inappropriate behavior, including threats, humiliation, or sexually suggestive content. However, because prompt engineering operates as a black-box approach, it lacks transparency, cannot consistently detect nuanced jailbreak attempts, and provides no clear rationale for its responses. These limitations highlight the need for a more interpretable, feature-based detection system - leading to the motivation for this study.

## 4 Results

**Table 3.** Performance Metrics of Detection Methods

| Detection Method | Accuracy | Precision | Recall | F1-Score | ROC-AUC |
|---|---|---|---|---|---|
| DT | 0.9306 | 0.9556 | 0.9072 | 0.9307 | 0.9118 |
| DT3 | 0.9284 | 0.9322 | 0.9283 | 0.9302 | 0.9798 |
| FDT | **0.9479** | 0.9532 | 0.9451 | **0.9492** | 0.9834 |
| GF | 0.9436 | 0.9253 | **0.9654** | 0.9449 | **0.9838** |
| RF | 0.9436 | 0.9567 | 0.9325 | 0.9444 | 0.9722 |
| LGBM | 0.9393 | 0.9524 | 0.9283 | 0.9402 | 0.9811 |
| XGBoost | 0.9436 | 0.9567 | 0.9325 | 0.9444 | 0.9776 |
| LR | 0.9458 | **0.9569** | 0.9367 | 0.9467 | 0.9818 |
| NN | 0.9393 | 0.9392 | 0.9392 | 0.9392 | - |
| PE | 0.8119 | 0.7429 | 0.9614 | 0.8381 | - |

The models were trained using an 80/20 train-test split to evaluate their ability to detect jailbreak prompts. Table 3 presents the performance metrics of all methods evaluated in this study, including Accuracy, Precision, Recall, F1-Score, and Area Under the Curve (AUC).

Key Observations:

- Fuzzy-based models (FDT and GF) outperformed all other approaches across multiple metrics. This may be attributed to the fuzzy logic framework, which better captures the uncertainty and gradation inherent in human language annotations. Among them, FDT showed slightly better accuracy and F1-score than GF, indicating a stronger balance between precision and recall.

- Surprisingly, despite its simplicity and assumption of linear decision boundaries, LR emerged as the second-best performing model overall, achieving the highest precision (0.9569) and competitive scores across all other metrics. This suggests that the feature set is linearly informative, allowing even simple models to perform exceptionally well.

- Decision Trees (DT and DT3) performed nearly as well as more complex black-box models. While the default decision tree achieved higher accuracy and precision, the shallower tree demonstrated better recall and the highest AUC. The structure of DT3 (Figure 1) also reveals that jailbreaking is strongly associated with distraction and irrelevance, supporting the hypothesis that annotators followed a guideline emphasizing alignment with the educational purpose of the simulation and the AI's role as a virtual patient.

- Prompt Engineering performed the worst among all methods, highlighting its limitations as a standalone strategy for jailbreak prevention.
  - False positives occurred when students used language such as "pee" or "poop" - phrasing that, while informal, can be medically relevant in context. PE failed in these cases due to its rigid enforcement of professionalism, lacking nuance in interpreting acceptable clinical language.
  - False negatives were more concerning. For instance, students could order many unnecessary tests - a subtle yet intentional deviation meant to confuse or test the system - but PE failed to flag this behavior. Moreover, even when explicitly instructed not to provide direct answers, the LLM occasionally generated diagnoses when prompted, or complied when instructed to ignore the guidelines. These behaviors represent unpredictable failure modes, showing that PE lacks robustness and flexibility when handling subtle or adversarial inputs.

The results suggest that the four extracted linguistic variables - Professionalism, Medical Relevance, Ethical Behavior, and Contextual Distraction - effectively capture patterns associated with jailbreak behavior in the dataset. Even basic classification models performed well using these features, indicating that they provide a strong, interpretable foundation for detecting deviations from the intended educational interaction.

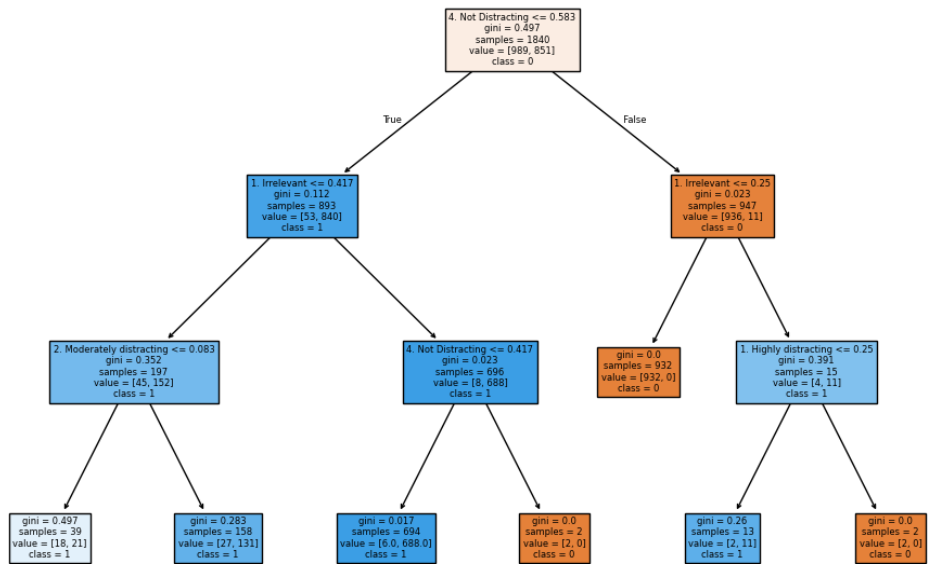

**Fig. 1.** Decision Tree (Max Depth = 3) for Jailbreak Classification Using Extracted Features.

## 5    Conclusion

This study explored jailbreak detection in an educational LLM system through a two-stage approach: manual data annotation based on four interpretable linguistic variables, and the development of predictive models using those features. Among all methods evaluated, the Fuzzy Decision Tree achieved the highest overall performance, demonstrating that even relatively simple models can effectively detect jailbreak behavior when grounded in meaningful annotations. The results indicate that feature-based prediction offers strong, interpretable performance, while prompt engineering alone is not sufficient for reliably detecting or preventing jailbreaks - particularly when users engage in creative or adversarial prompting strategies. However, a key limitation remains: the current approach does not address real-time detection, where prompts must be annotated and interpreted accurately as they occur. Looking forward, future work could explore hybrid models that combine the strengths of both approaches. Prompt-based methods excel at capturing dynamic context and adapting to novel behavior beyond the training data, while rule-based or feature-driven systems offer stability, robustness, and the ability to be updated when new jailbreak strategies emerge. Rather than treating jailbreak detection as a binary task, a spectrum-based approach could allow for uncertainty: for example, rule-based models could trigger deeper prompt-based investigation when confidence is low. This hybrid framework could lead to more flexible, transparent, and trustworthy LLM systems in sensitive applications like medical education.